\documentclass[10pt,a4paper]{article}
\usepackage{f1000_styles}

\usepackage[numbers]{natbib}
\usepackage{graphicx}
\usepackage{subfigure}
\usepackage{todonotes}
\usepackage{algorithm}
\usepackage{algorithmic}
\usepackage{float}

\begin{document}
\pagestyle{fancy}

\title{Optimizing Energy Efficiency in Metro Systems Under Uncertainty Disturbances Using Reinforcement Learning}
\author[1]{Haiqin Xie}
\author[1]{Cheng Wang}
\author[2]{Shicheng Li}
\author[2]{Yue Zhang}
\author[3]{Shanshan Wang}
\affil[1,2,3]{Haier Digital Technology (Shanghai) Co., Ltd.,Shanghai, 200000, China}

\maketitle
\thispagestyle{fancy}

\
\\
\\
\begin{abstract}
In the realm of urban transportation, metro systems serve as crucial and sustainable means of public transit. However, their substantial energy consumption poses a challenge to the goal of sustainability. Disturbances such as delays and passenger flow changes can further exacerbate this issue by negatively affecting energy efficiency in metro systems. To tackle this problem, we propose a policy-based reinforcement learning approach that reschedules the metro timetable and optimizes energy efficiency in metro systems under disturbances by adjusting the dwell time and cruise speed of trains. Our experiments conducted in a simulation environment demonstrate the superiority of our method over baseline methods, achieving a traction energy consumption reduction of up to 10.9\% and an increase in regenerative braking energy utilization of up to 47.9\%. This study provides an effective solution to the energy-saving problem of urban rail transit.
\end{abstract}

\section*{\color{primaryColour}Keywords}

\textbf{Metro System, Uncertainty Disturbance, Energy Efficient, Timetable Rescheduling, Reinforcement Learning}

\clearpage
\pagestyle{fancy}

\section{Introduction}

Metro rail transit has become an essential part of urban transportation infrastructure in large and medium--sized cities worldwide. As the number of subway projects continues to grow, and the total length of subway lines increases, the energy demand for these systems has surged, leading to higher energy costs and environmental impacts. Therefore, optimizing energy efficiency in metro systems has become an urgent issue to ensure sustainable and cost--effective operations.
	
    \
	
Building an Energy-Efficient Train Timetabling (EETT) \cite{Scheepmaker2017} is an offline optimal method that can reduce running power consumption in metro rail transit systems by setting fixed departure and arrival times for trains. By avoiding unnecessary train operations during low--demand periods, EETTs can help conserve energy. However, they are vulnerable to uncertainties and disturbances that can considerably reduce their effectiveness. Sudden changes in passenger demand, track congestion, and weather conditions can disrupt scheduled train operations, leading to increased energy consumption and delays. Moreover, unforeseen incidents such as equipment failures or accidents can cause disruptions that require immediate adjustments to the train schedule, making EETTs inefficient. Therefore, it is critical to employ advanced techniques that can adapt to real--time variations and disturbances to improve the energy efficiency of metro rail transit systems.
	
    \
	
Train timetable rescheduling (TTR) \cite{Binder2016} is a flexible and adaptable approach to train scheduling that can handle unforeseen events and enhance the energy efficiency of metro rail transit systems. TTR can be utilized to modify train schedules in response to sudden changes in passenger demand, track congestion, weather conditions, and unforeseen incidents such as equipment failures or accidents. Unlike EETTs, which have fixed schedules, TTR can adjust the train timetable in real--time, making it more responsive to changes in passenger demand and traffic conditions. Consequently, TTR can improve the energy efficiency of metro rail transit systems by reducing energy consumption and minimizing delays. By employing advanced control systems and planning tools that can adapt to real--time variations and disturbances, TTR can ensure that train services remain efficient and responsive to the needs of passengers and stakeholders.
	
    \	
	
In this paper, we propose the use of Proximal Policy Optimization (PPO) \cite{Schulman2017}, a policy-based reinforcement learning algorithm, to address the problem of uncertain disturbances in metro rail transit systems and enable efficient TTR for metro operation. PPO is known for its robustness, scalability, efficiency, flexibility, and state--of--the--art performance, making it well-suited for optimizing complex systems based on real-time data. By receiving real--time data on other trains' information, the PPO algorithm can dynamically adjust the train schedule to minimize energy consumption. Our approach focuses on dynamically adjusting the metro train's dwell time and cruise speed to optimize energy efficiency, even in the presence of uncertainties and disturbances. This proposed approach has the potential to significantly reduce energy consumption in metro rail transit systems, making them more sustainable, efficient, and cost-effective.
	
    \
	
In the following sections, we first review related works on optimizing energy efficiency in metro train systems. Then, we introduce our modeling approach to simulate the train operation and energy consumption. Next, we describe our simulation environment and reinforcement learning method based on the PPO algorithm to optimize train control decisions and reduce energy consumption. We present and analyze the experimental results to demonstrate the effectiveness of our approach in terms of energy savings. Finally, we summarize our contributions.

\section{Related Works}
\label{sec: Related Works}
The energy consumption of metro trains is greatly influenced by speed profiles and train schedules, which are crucial factors that affect the power required for train operations. Previous studies have focused on two primary areas for optimizing energy consumption: speed profile optimization and timetable optimization. To optimize speed profiles, researchers have employed various methods such as analysis and numerical techniques. On the other hand, optimizing train schedules requires the use of integer programming (IP) or genetic algorithms (GA). Recently, reinforcement learning has emerged as a popular method for optimizing train schedules.

\subsection{Speed Profiles Optimization}
In 1968, Ishikawa \cite{ichikawa1968application} proposed an optimal control model for train speed curves, simplifying train resistance as a linear function of speed and dividing the control process into four stages: maximum acceleration, cruising, coasting, and maximum braking. Later, Strobel \cite{strobel1974contribution} omitted the cruise process for urban rail transit and introduced ramp gradient. Milroy \cite{milroy1980aspects} then developed a metro model with an optimal driving control strategy that includes only maximum acceleration, coasting, and maximum braking. For gentle slope routes, Khmelnitsky \cite{khmelnitsky2000optimal} verified that there is only a stable optimal cruising and coasting strategy. T. Albrecht and Oettich \cite{albrecht2002new} used Simulink simulation software to solve the optimal single train speed curve, which saved 15\% to 20\% energy in experiments. Aradi \cite{aradi2013predictive} used predictive optimization with multiple objectives, including minimizing energy consumption and maximizing on-time arrival at the terminal.
	
\subsection{Timetable Optimization}
Ghoseiri \cite{ghoseiri2004multi} developed a nonlinear mathematical programming problem to select an optimal metro timetable that minimizes passenger travel time and fuel consumption. Chen \cite{chen2005optimization} used a genetic algorithm to optimize metro timetables while avoiding simultaneous acceleration of multiple trains, which can cause peak instantaneous power to exceed the threshold. Yong \cite{yong2011two} created a two-layer iterative optimization model that reduced energy consumption by 19.1\% by modeling the relationship between metro timetables and energy consumption. Kim \cite{kim2011model} employed mixed-integer programming to minimize peak energy consumption while maximizing regenerative energy. Gong's \cite{Gong2014} Integrated-Energy-Efficient Operation Methodology (EOM) reduces travel time and recovers to the original EETT using a genetic algorithm optimizer. Ning \cite{ning2019deep} used deep reinforcement learning to reduce the total delay by 46.38\% compared to the First-Come-First-Served method. Yang \cite{yang2019real} proposed RTTRM, which combines GA and DNN to optimize metro timetables and reduce energy consumption. Liao \cite{liao2021deep} proposed using deep reinforcement learning methods instead of GA and achieved a 5.11\% improvement.

\section{Modeling}
\label{sec: Modeling}
In this section, we present a comprehensive modeling approach for the metro network, which involves time modeling, mechanical modeling, and energy modeling. This multi-faceted approach allows us to capture the complex dynamics of the system, including train operations, power consumption, and regenerative braking energy. By considering these different aspects of the metro network, we are able to develop a more accurate and robust simulation framework, which can be used to evaluate the performance of reinforcement learning algorithms and optimize the operation of the system.

\subsection{Time Model}
The time model of a metro train specifies the departure instants of the different trains from the different platforms \cite{van1991traffic}. It consists of three key parameters: departure time, travel time, and dwell time. The departure time $t_{de}$ is the time at which a train leaves the station, while the travel time $t_{tr}$ is the duration it takes for the train to travel from one station to the next. The dwell time $t_{dw}$ is the amount of time a train spends at a station before departing for the next one. The superscript of $t$ includes two variables, indicating the train and station index, respectively. Additionally, $t_{h}$ represents the interval time between successive metro train departures. The departure instant from the first station for each metro train can be calculated as:

    \begin{equation}
        t^{m,1}_{de} = (m-1)t_{h}
    \end{equation}
    
The departure time $t_{de}$ of the $m^{th}$ metro train at the $n^{th}$ station can be calculated using the following equation. 
    
    \begin{equation}
        t^{m,n}_{de} = t^{m,1}_{de} + \sum_{i=2}^{n}(t^{m,i}_{tr}+t^{m,i}_{dw})
    \end{equation}	
    
In the event of a disturbance at metro station No. $n$, the departure time of metro train No. $m$ from that station will be affected. The departure time function with disturbance, which includes the variable $t_{\epsilon}$, is shown in the equation below.	

    \begin{equation}
        t^{m,n}_{de} = t^{m,1}_{de} + \sum_{i=2}^{n}(t^{m,i}_{tr}+t^{m,i}_{dw})+ t_{\epsilon}
    \end{equation}
    
The total time for each metro train can be calculated using the following equation. However, it is important to note that since the last station is a terminal, the dwell time and disturbance time are not included in the total time calculation. Then, the total departure instant will be represented by:

    \begin{equation}
        t^{M,N}_{total} = t^{m,1}_{de} + \sum_{i=2}^{N-1}(t^{M,i}_{tr}+t^{M,i}_{dw} + t_{\epsilon}) + t^{M,N}_{tr}
    \end{equation}
		
where $M$ is the total number of metro trains and $N$ is the total number of metro stations.

\subsection{Mechanical Model}
		
During the acceleration phase, only the traction force is present, while the braking force is zero. As the speed of the metro train increases from zero to $v_1$, the electric motor maintains a constant torque output at its rated power. Beyond $v_1$, the power output remains constant while the torque becomes inversely proportional to the speed of the train. Therefore, the traction force can be described as follows:

    \begin{equation}
        T_T(v^m) =
        \left\{
            \begin{aligned}
                &\:\:\:\:\:\:\:p_1		&0<v^m<v_1	\\
                &\frac{q_1}{v^m + p_2}	&v^m>v_1
            \end{aligned}
        \right.
    \end{equation}	

where the $T$ represents the torque force, and the $v^m$ represents the speed of the No. $m$ metro train. Parameters $p_1$, $p_2$, and $q_1$ are given in relation to the traction force of the metro trains.

    \

During the cruising phase, when the traction and braking forces are zero, the acceleration of the train is solely determined by the resistance force, which can be expressed as $F_R(v^m)$ and is governed by the Davis equation \cite{davis1926tractive} shown below. The resistance force acts in opposition to the train's movement, and the resulting acceleration is given by $a = F_R(v^m) - G\sin\theta(x^m)$, where $G\sin\theta(x^m)$ is the component of gravity in the direction of movement, related to the slope of the track $\theta(x^m)$:

    \begin{equation}
        T_R(v^m) = \lambda_{1}(v^m)^2 + \lambda_{2}v^m + \lambda_{3}
    \end{equation}
			
The parameters $\lambda_1$, $\lambda_2$, and $\lambda_3$ are dependent on various environmental conditions, as well as the shape of the trains.

    \

During the braking phase, the braking torque force exhibits an inversely proportional relationship with the speed as long as the speed is greater than $v_2$, and the braking torque remains constant. As the speed decreases and becomes equal to or less than $v_2$, the braking torque becomes fixed. The braking force can be described as follows:

    \begin{equation}
        T_B(v^m) =
        \left\{
        \begin{aligned}
            &\:\:\:\:\:\:\:p_3		&0<v^m<v_2	\\
            &\frac{q_2}{v^m + p_4}	&v^m>v_2
        \end{aligned}
        \right.			
    \end{equation}
	
\subsection{Energy Model}
In a metro train network, the metro trains have three states: acceleration, cruising, and braking \cite{howlett1996optimal}. An accelerated metro train can utilize the regenerative energy generated by a simultaneously braking metro train. To estimate the essential traction power as well as a reproduced braking energy, the total power application of a network is obtained from the following \cite{kuppusamy2020deep}

    \

$P_T^M$ is the traction power of the No. $M$ metro train and is defined as:

    \begin{equation}
        P_T^M(t) = \left\{
            \begin{aligned}
                &\displaystyle\frac{T_Tv^m}{\beta_1}  &T_T>0\\
                &\:\:\:\:\:0 &T_T\leq0
            \end{aligned}
        \right.
    \end{equation}

where $\beta_1$ is the conversion efficiency from electrical energy to mechanical energy. Therefore, the overall time traction energy is defined as:
		
    \begin{equation}
        E_T = \int_{0}^{T_s}\sum_{M=1}^{M}P_T^M(t)dt
    \end{equation}

$P_B^M$ is the braking power of the No. $M$ metro train, which is defined as:

    \begin{equation}
        P_B^M(t) = \left\{
            \begin{aligned}
                &\:\:\:\:\:\:\:\:\:\:\:\:0	& T_B\geq0	\\
                &-T_Bv^m\beta_2 & T_B<0
            \end{aligned}
        \right.
    \end{equation}

where $\beta_2$ is the conversion efficiency from mechanical energy to electrical energy.  

    \
		
At the time $t$, if overall traction power is greater than overall braking power, the truly used power would be $\beta_3\sum_{M=1}^{M}P_B^M(t)$, where $\beta_3$ is the conversion efficiency from electrical energy to mechanical energy. On the contrary, if braking power is greater than traction power, then the truly used power would be $\sum_{M=1}^{M}P_T^M(t)$. The excess part of breaking power would be saved in the metro's battery or sent to the energy storage device or converted into heat energy by resistance. Hence, the regenerative power at time $t$ would be:

    \begin{equation}
        P_R(t, \epsilon) = min(\sum_{M=1}^{M}P_T^M(t), \beta_3\sum_{M=1}^{M}P_B^M(t))
    \end{equation}

        \

    \begin{equation}
        E_R(\epsilon) = \int_{0}^{T_s}P_R(t,\epsilon)dt
    \end{equation}

The total net traction energy $E_{total}$ of the trains from the starting station to the terminal can be defined as:

    \begin{equation}
        E_{total} =E_T - E_R(\epsilon)
    \end{equation}	

\section{Method}
In this section, we present the rationale behind our method. Our approach is based on the principles of reinforcement learning, which involves training agents to make decisions by learning from the rewards and penalties received from the environment. By applying reinforcement learning to the problem of metro system energy optimization, we aim to develop a solution that is both adaptive and efficient, capable of adapting to changing conditions while maintaining optimal energy usage.

\subsection{Markov Decision Process Formulation}
Markov Decision Process (MDP) is a mathematical framework that is widely used in RL to model decision-making problems under uncertainty. It is a tuple of seven elements \cite{thomas2007markov}: state space($S$), action space($A$), initial state distribution($\mu_0$), transition dynamics($T$), reward function($r$), discount factor($\gamma$) and horizon($H$). In an MDP, an agent interacts with the environment by selecting actions that influence the state transition probabilities and immediate rewards. The agent's goal is to learn a policy that maximizes the expected cumulative reward over time. In this paper, we use MDP as the mathematical model to formalize our problem of optimizing the network's energy consumption.
	
    \begin{figure}[H]
        \centering
        \includegraphics[width=0.4\textwidth]{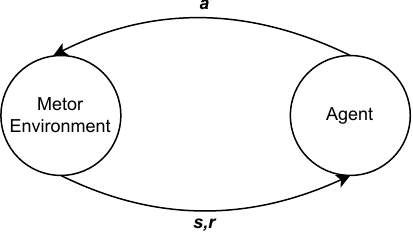}
        \caption{Environment and Agent}
        \label{fig: Environment and Agent}
    \end{figure}
	
\textbf{Environment and Agent.} 
The environment in this study is a simulated metro train network, where the agent makes decisions based on the state of the environment. After a disturbance occurs, the agent observes other metro trains' information and outputs the next cruise speed and the next station dwell time. The agent only observes the states at the moment it leaves the station to make a decision. The agent aims to optimize energy consumption by maximizing the use of regenerative braking energy while still meeting the schedule requirements.
		
    \
	
\textbf{States.}	
The states are the fundamental elements of the MDP that the agent uses to make decisions. In our case, the states are derived from the observations of the metro train network. The selection of appropriate state observations is a critical step that directly impacts the accuracy of the agent's decision-making. The dimensional of the state space is determined by the number of metro trains and the number of features. As the dimensional of the state space increases, the action network's complexity rises, leading to extended training time and potentially resulting in over-fitting of the training data. However, a more intricate network structure can enable the agent to gain a deeper understanding of the environment, reducing the risk of network under-fitting. To strike a balance between complexity and accuracy, we employ eight state observation features, namely the historical acceleration time point, historical constant speed time point, historical deceleration time point, historical dwell time point, current location, distance to the next station, train direction, and whether the train has reached the terminal.
		
    \
	
\textbf{Actions.}
The agent's decision-making is restricted to the moment of train departure and only pertains to determining the next station's cruising speed and dwell time. This design ensures that the dimensional of the agent's policy output is consistently two, regardless of the number of trains or stations in the test segment. The agent outputs actions within the range of $[-1, 1]$, which are then mapped to the value inside the bounds of dwell time and cruising speed. This mapping ensures that the agent's decisions fall within the practical limits of the system, and the actions can be applied to the subway trains.

    \

\textbf{State transition.}
Once the agent selects the actions $a_t$ at time $t$, the metro train network environment runs continuously until a new decision is required for one of the metro trains at time $t^{\prime}$. The environment then transitions to the next state $s_{t^{\prime}}$ based on the chosen actions. The environment will simulate and execute the train's movement based on the modeling outlined in Section \ref{sec: Modeling} with additional disturbance time added when stopping at the station. The new state is observed, and the agent generates new actions $a_{t^{\prime}}$ based on the updated information. This process repeats, with the agent making decisions and the environment simulating the results, until the end of the simulation.

    \begin{figure}[H]
        \centering
        \includegraphics[width=0.8\textwidth]{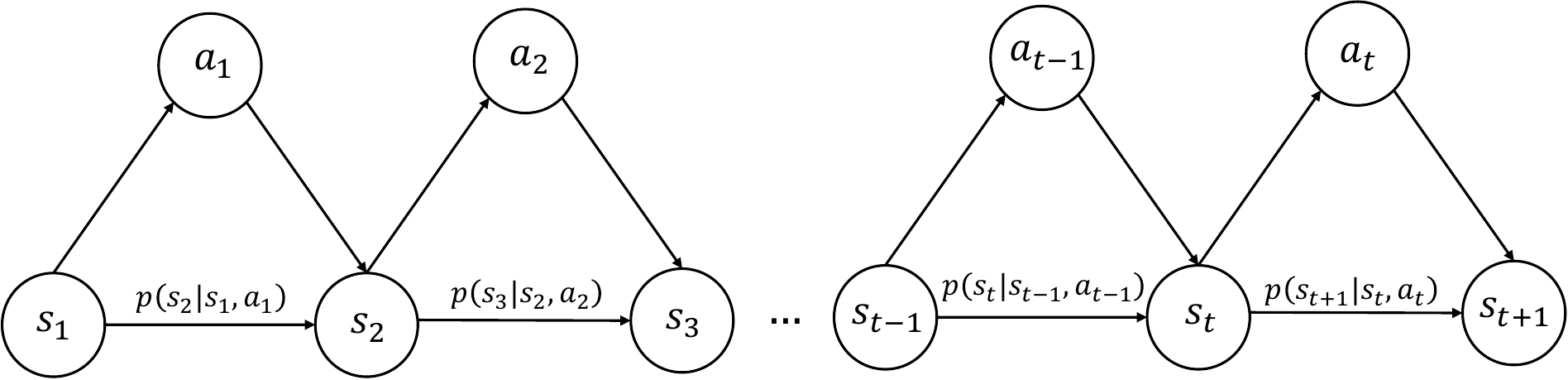}
        \caption{States Transition}
        \label{fig: states transition}
    \end{figure}
	
\textbf{Rewards.}	
The reward was designed to maximize the utilization of regenerative braking energy in the metro train network. To achieve this objective, we developed a reward function that calculates the overlap time between the braking time and the traction time for all trains in the system. The reward function considers the actions selected by the agent and compares them to the previous overlap time that was calculated in the previous step. By doing so, we can accurately determine the utilization of regenerative braking energy that has been utilized since the last step, which becomes the step reward. This reward function incentivizes the agent to optimize the use of regenerative braking energy and encourages efficient operation of the metro system. 

\subsection{Framework}
\begin{figure}[H]
        \centering
        \includegraphics[width=0.5\textwidth]{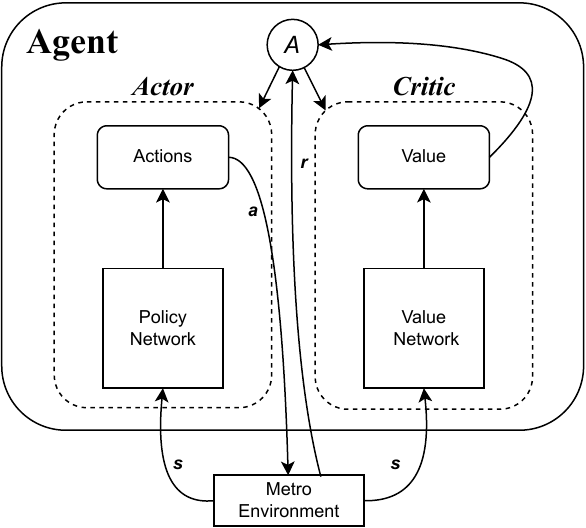}
        \caption{Actor-Critic Framework of Agent}
        \label{fig: Actor-Critic Framework}
    \end{figure}

Our framework is based on the actor--critic method, which combines both value-based and policy-based approaches. The actor-critic method has two components: an actor, which learns the optimal policy, and a critic, which learns the value function of the policy. Just as the figure \ref{fig: Actor-Critic Framework} shows, the agent consists of two neural networks, one for the actor and one for the critic, that map input data to output data. The actor chooses optimal decisions based on its policy, and the critic provides feedback to the actor on the quality of its current policy. This feedback helps the actor update its policy to improve its performance over time, which will be discussed in subsection \ref{subsection: Policy Update}. 

\textbf{Actor.} The actor in our framework is designed to learn the optimal policy through iterations. It consists of a policy network ($\pi_{\theta}$) that maps the received states from the metro environment to decision actions for the metros to execute. Policy network can also be viewed as a probability distribution which can be mathematically represented as:

    \begin{equation}
        a_t = \pi_{\theta}(s_t)	
    \end{equation}
    
During each episode($\tau$), the policy is calculated using the following function:

    \begin{equation}
        \pi_{\theta}(\tau) = P(s_1)\prod_{t=1}^{T}\pi_{\theta}(a_t|s_t)P(s_{t+1}|s_t,a_t)
    \end{equation}
    
Here, $P$ represents the probability of state transition, and $\theta$ represents the weight set of the policy network of the actor. 


    \

\textbf{Critic.} The critic in our framework is designed to estimate the value function of the policy. It takes in the state information from the metro environment and provides feedback to the actor on how good its current policy is. The value function is estimated using a value network ($V_{\phi}$) with weights represented by $\phi$. The value network estimates the state-value function $V_{\pi}(s_t)$, which represents the expected sum of future rewards from state $s_t$ under the current policy $\pi$.

    \

The state-value function can be estimated by the following Bellman equation:
	
    \begin{equation}
    V_{\pi}(s) = \mathbb{E}_{\pi}[r+\gamma V_{\pi}(s^{\prime})],
    \end{equation}
	
where $r$ is the immediate reward obtained at state $s$, $s^{\prime}$ is the next state obtained by taking action $a$ from state $s$, and $\gamma$ is a discount factor that determines the weight of future rewards.

    \
	
\textbf{Advantage Function.} The advantage function is a key component as it provides a measure of how good a particular action is in a given state, which can be used to update the policy to improve performance. It can be calculated using the following expression:

    \begin{equation}
        A(s_t, a_t) = r(s_t, a_t) + V(s_{t+1}) - V(s_t)		
    \end{equation} 

where $r(s_t, a_t)$ is the reward received for taking action $a_t$ in state $s_t$, $V(s_{t+1})$ is the expected value of the next state $s_{t+1}$ under the current policy, and $V(s_t)$ is the expected value of the current state $s_t$ under the current policy. The advantage function essentially measures the difference between the expected reward of taking a particular action in a given state and the expected reward of following the current policy in that state. If the advantage is positive, then the action is better than the current policy, and if it is negative, then the action is worse than the current policy.


	
\subsection{Policy Update}
\label{subsection: Policy Update}
The policy update in our framework is based on the PPO. The objective of PPO is to maximize the expected cumulative reward while constraining the policy update to a neighborhood around the current policy. This constraint is called the trust region, and it ensures policy stability by preventing large policy swings that could destabilize the learning process. 
	



    \

\textbf{Loss Functions.} The PPO algorithm has three loss functions: the clipped surrogate objective function, the value function loss, and the entropy loss. 

    \

The clipped surrogate objective function, which is a modification of the policy gradient loss, is given by:

    \begin{equation}
        L^{CLIP}(\theta) = \hat{\mathbb{E}}_t\Big[\min\big(p_t(\theta)\,\hat{A}_t,  \text{clip}(p_t(\theta), 1 - \epsilon, 1 + \epsilon)\,\hat{A}_t\big)\Big] 
    \end{equation}
 
where $\hat{\mathbb{E}}_t$ denotes an empirical estimate of the expectation, $p_t(\theta) = \frac{\pi_{\theta}(a_t|s_t)}{\pi_{\theta_{old}}(a_t|s_t)}$ is the probability ratio, $\hat{A}_t$ is an estimator of the advantage function, and $\epsilon$ is a hyper-parameter controlling the size of the trust region.

		\

The value function loss is given by:

    \begin{equation}
    L^{VF}(\phi) = \hat{\mathbb{E}}_t\Big[(V_{\phi}(s_t) - V_t^{target})^2\Big]
    \end{equation}
    
where $V_\phi(s_t)$ is the predicted value function, and $V_t^{target}$ is an estimate of the true value function.
	
		\

The entropy loss is given by:

    \begin{equation}
    L^{S}(\pi_{\theta}) = \hat{\mathbb{E}}_t\Big[\pi_{\theta}(a_t|s_t)\log\pi_\theta(a_t|s_t)\Big]
    \end{equation}
    
which encourages exploration by penalizing overly deterministic policies.
    
    \
    
\textbf{Algorithm.} The achievement of the PPO algorithm for metro decision-making can be shown through the pseudocode presented in Algorithm \ref{alg:ppo}. The input parameters of the algorithm include the policy parameterization $\pi_{\theta}(a|s)$ and value function parameterization $V_{\phi}(s)$, and the output is the updated policy parameters $\theta$. In each step, the policy is run in the metro simulation environment and the gained experience is saved in a buffer. Advantages are then computed, and the surrogate loss, value loss, and entropy loss are calculated. The total loss is obtained by adding these three losses and updating the parameters $\theta$ and $\phi$ by minimizing the total loss using Adam. 

    \begin{algorithm}[H]
        \caption{Proximal Policy Optimization (PPO) for Metro Decision-Making}
        \label{alg:ppo}
        \renewcommand{\algorithmicrequire}{\textbf{Input:}}
        \renewcommand{\algorithmicensure}{\textbf{Output:}}
        \begin{algorithmic}[1]
            \REQUIRE
            Policy parameterization $\pi_{\theta}(a|s)$ and Value function parameterization $V_{\phi}(s)$
            \ENSURE
            Updated policy parameters $\theta$
            \FOR{each iteration}
                \FOR{each step of episode}
                    \STATE Run Policy $\pi_{\theta}(a|s)$ in metro simulation environment
                    \STATE Save experience in buffer
                \ENDFOR    
                \STATE Compute advantages $A$
                \STATE Compute surrogate loss $L^{\text{CLIP}}(\theta)$, value loss $L^{\text{VF}}(\phi)$ and entropy loss $L^{\text{S}}[\pi_{\theta}(a|s)]$
                \STATE Compute total loss $L(\theta, \phi) = L^{\text{CLIP}}(\theta) + c_1 L^{\text{VF}}(\phi) - c_2 L^{\text{S}}[\pi_{\theta}(a|s)]$
                \STATE Update $\theta$ and $\phi$ by minimizing $L(\theta, \phi)$ using Adam
            \ENDFOR
        \end{algorithmic}
    \end{algorithm}			
		
\section{Experiment}
In this section, we present the experimental results that demonstrate the effectiveness of our approach in reducing energy consumption. The experiments were conducted on a realistic simulation of a metro system under uncertainty disturbance, and the results showed that the PPO algorithm was highly effective in reducing energy consumption. Furthermore, the PPO algorithm was able to adapt to different levels of uncertainty disturbance, showcasing its robustness and versatility.
		
    \

\textbf{Datasets and Baseline.} We constructed our simulation environment using the Xiamen line 1 dataset, which comprises 24 stations and their respective distances, as well as the dwell time. Additionally, the dataset includes inferred cruise speed values based on distance and running time, which ensured the realism of our experiments. Prior to running our simulations, we meticulously validated the dataset and performed any required pre-processing steps to ensure the accuracy and reliability of our results.

	\begin{table}[H]
		\caption{Xiamen Line 1 Dataset}
		\centering
		\begin{tabledata}{l^c^c^c}
			\header Departure - Arrival & Distance(km) & Cruise Speed(km/h) & Dwell time(s)\\
			\row Zhenghailuzhan -- Zhongshangongyuanzhan & 0.89 & 68.4 & 25 \\
			\row Zhongshangongyuanzhan -- Jiangjuncizhan & 1.16 & 58.9 & 25 \\
			\row Jiangjuncizhan -- Wenzaozhan & 0.92 & 65.4 & 25 \\
			\row Wenzaozhan -- Hubingdongluzhan & 1.39 & 80.0 & 25 \\
			\row Hubingdongluzhan -- Lianbanzhan & 0.80 & 55.2 & 30 \\
			\row Lianbanzhan -- Lianhuahukouzhan & 1.02 & 77.5 & 25 \\
			\row Lianhuahukouzhan -- Lvcuozhan & 1.04 & 80.0 & 40 \\
			\row Lvcuozhan -- Wushipuzhan & 0.95 & 69.4 & 25 \\
			\row Wushipuzhan -- Tangbianzhan & 0.91 & 64.3 & 25 \\
			\row Tangbianzhan -- Huojuyuanzhan & 0.69 & 49.3 & 40 \\
			\row Huojuyuanzhan --Dianqianzhan & 1.62 & 80.0 & 25 \\
			\row Dianqianzhan -- Gaoqizhan & 1.48 & 80.0 & 25 \\
			\row Gaoqizhan -- Jimeixuecunzhan & 3.94 & 80.0 & 25 \\
			\row Jimeixuecunzhan -- Yuanboyuanzhan & 2.66 & 80.0 & 25 \\
			\row Yuanboyuanzhan -- Xinglincunzhan & 0.93 & 74.5 & 35 \\
			\row Xinglincun -- Xingjinluzhan & 0.62 & 45.4 & 25 \\
			\row Xingjinluzhan -- Guanrenzhan & 1.60 & 80.0 & 30 \\
			\row Guanrenzhan -- Chengyiguangchangzhan & 1.32 & 80.0 & 25 \\
			\row Chengyiguangchangzhan -- Ruanjianyuanzhan & 1.43 & 80.0 & 25 \\
			\row Ruanjianyuanzhan -- Jimeidadaozhan & 1.24 & 80.0 & 25 \\
			\row Jimeidadaozhan -- Tianshuiluzhan & 1.26 & 72.5 & 35 \\
			\row Tianshuiluzhan -- Xiamengbeizhan & 1.65 & 80.0 & 35 \\
			\row Xiamengbeizhan -- Yanneibeiguangchangzhan & 0.86 & 78.5 & 35 
		\end{tabledata}
		\label{table: Metro Station Data}
	\end{table}
	
From the data presented in table \ref{table: Metro Station Data}, it is evident that the maximum speed allowed inside the stations is 80 km/h, which plays a crucial role in ensuring the safety and efficiency of the metro system. In order to simulate the real-world scenario, we used 20 metros in the simulation environment, where half of the metros moved in the same direction, while the other half moved in the opposite direction. This approach helped us to take into account the real-world scenario where multiple metros run on the same track, and some of them move in the opposite direction.

    \

To ensure the accuracy and reliability of our simulations, the baseline result of the utilization of regenerative braking energy is obtained by incorporating the data from the Xiamen line 1 dataset into our metro simulation environment construction and adding uncertainty disturbance to the simulation, which is called "No action" that shown in table \ref{table:comparision}. This baseline result serves as a benchmark for our experiments and provides valuable insights into the performance of the system under normal operating conditions.

	

		\

\textbf{Training Configurations.} Our model training and parameter configurations are critical in achieving the desired performance in our experiments. To accomplish this, we utilize a neural network to approximate the policy and value functions in the PPO algorithm. To ensure optimal performance, we carefully fine-tuned our training parameters, which are detailed in table \ref{tab:model-params}.	

	\begin{table}[H]
		\centering
		\caption{Parameters Configuration}
		\label{tab:model-params}
		\begin{tabledata}{l^c^c}
			\header \textbf{Hyperparameter} & \textbf{Value} \\ 
			\row Number of Steps & 480 \\                
			\row Batch Size & 64 \\
			\row Gamma & 0.99 \\
			\row Clip Range & 0.2 \\
			\row Value Function coefficient & 0.5 \\
			\row Entropy coefficient & 0.1 \\
			\row Learning rate & 3e-4 \\
		\end{tabledata}
	\end{table}

\begin{figure}[H]
		\centering
            \subfigure[Episode Reward Mean]{
		      \includegraphics[width=0.4\textwidth]{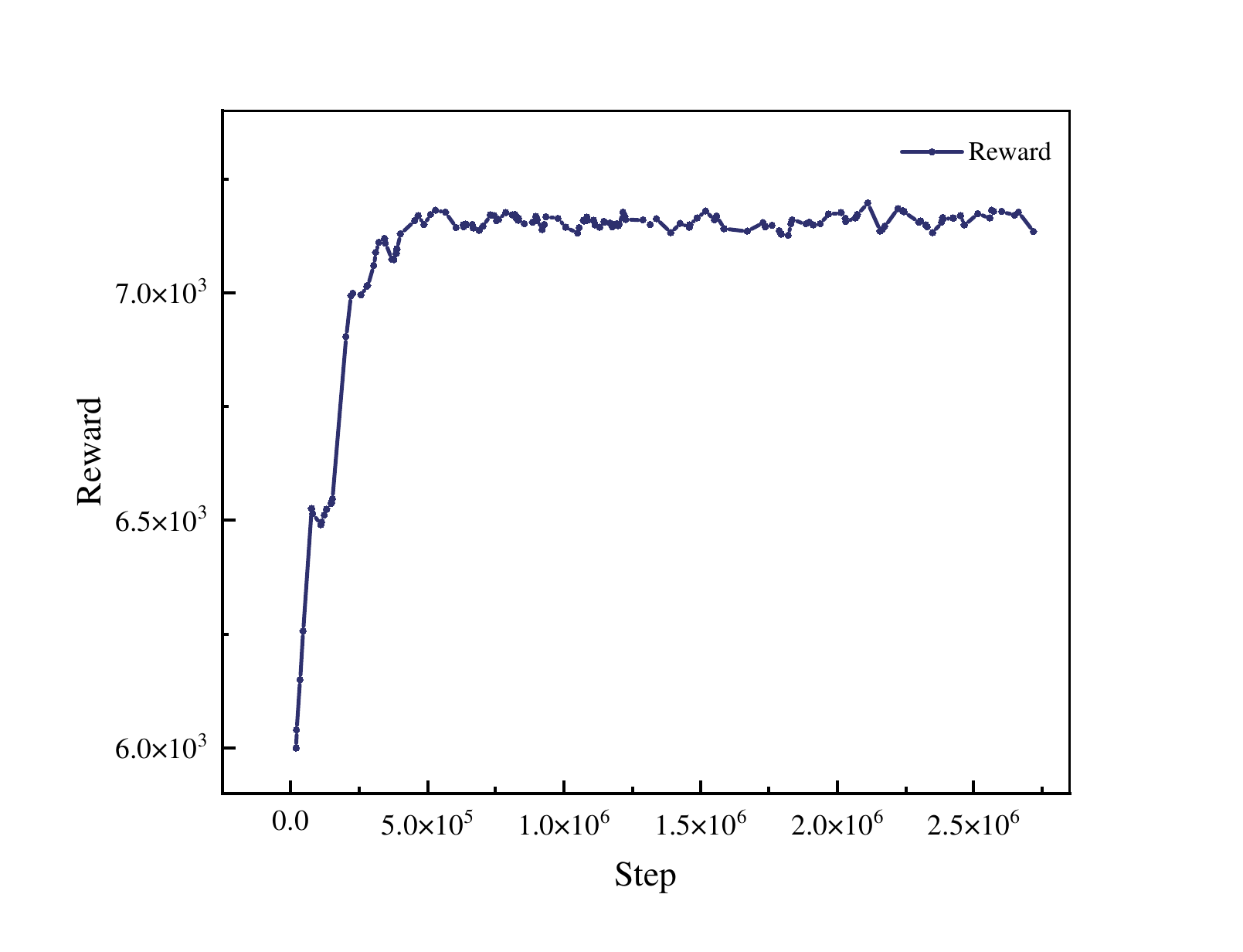}  
                \label{fig:ep_rew_mean}
            }
            \subfigure[Training FPS]{
                \includegraphics[width=0.4\textwidth]{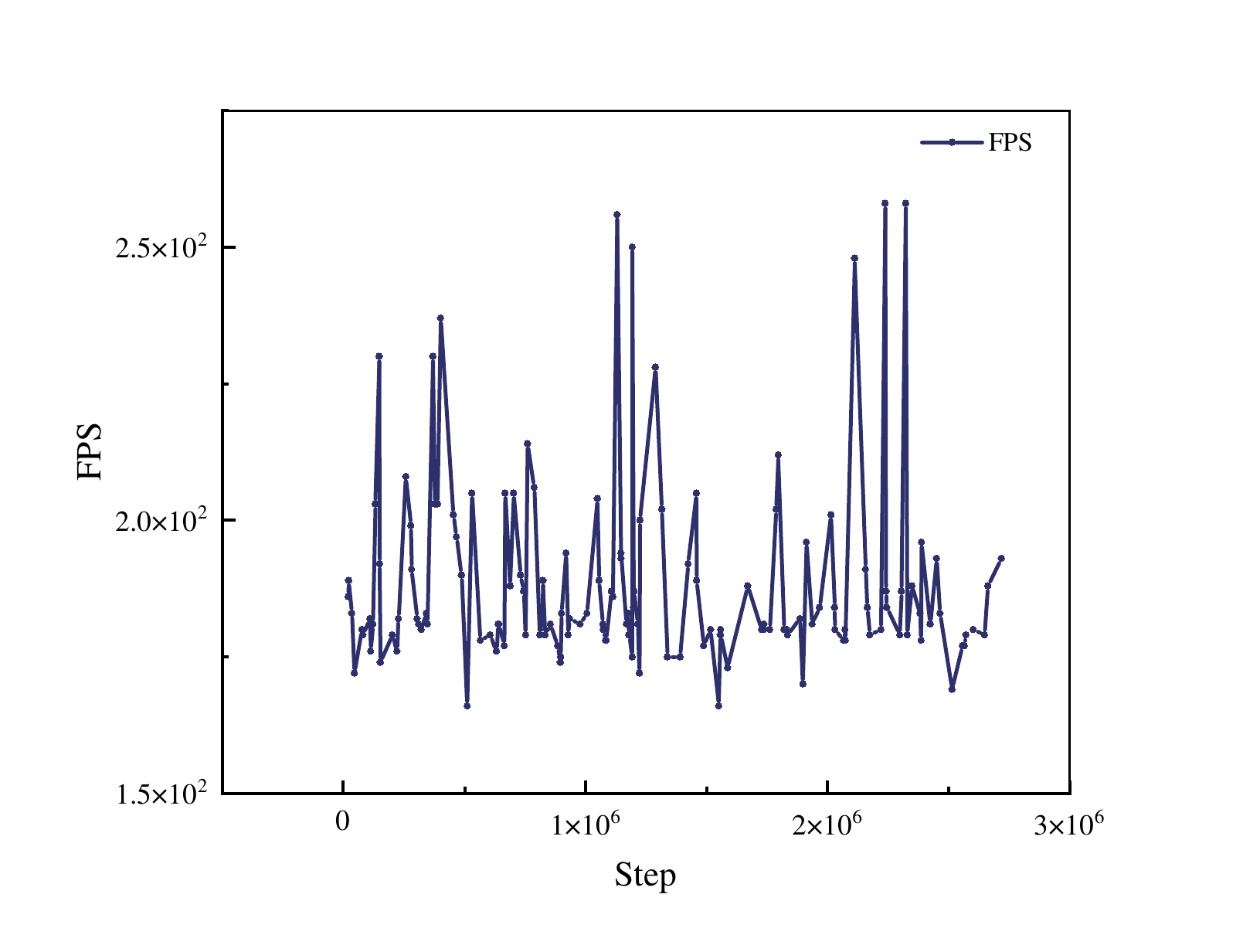}  
                \label{fig: fps}
            }
            \caption{Roll-out Result during Training}
            \label{fig: Training result}
	\end{figure}

\textbf{Results.} Figure \ref{fig:ep_rew_mean} shows the mean reward obtained by the agent during training as the number of steps increases, indicating that the agent learns to optimize regenerative power utilization. The graph shows a steady increase in the mean reward, and the reward line eventually converges to a stable value, indicating that the agent has learned the optimal policy. These results demonstrate the effectiveness of our approach in maximizing energy efficiency in metro systems. Figure \ref{fig: fps} displays the frames per second(FPS) during training, which shows that our approach is efficient and capable of handling large-scale simulation environments.

\begin{figure}[H]
		\centering
		\subfigure[Policy Gradient Loss]{
			\includegraphics[width=0.3\textwidth]{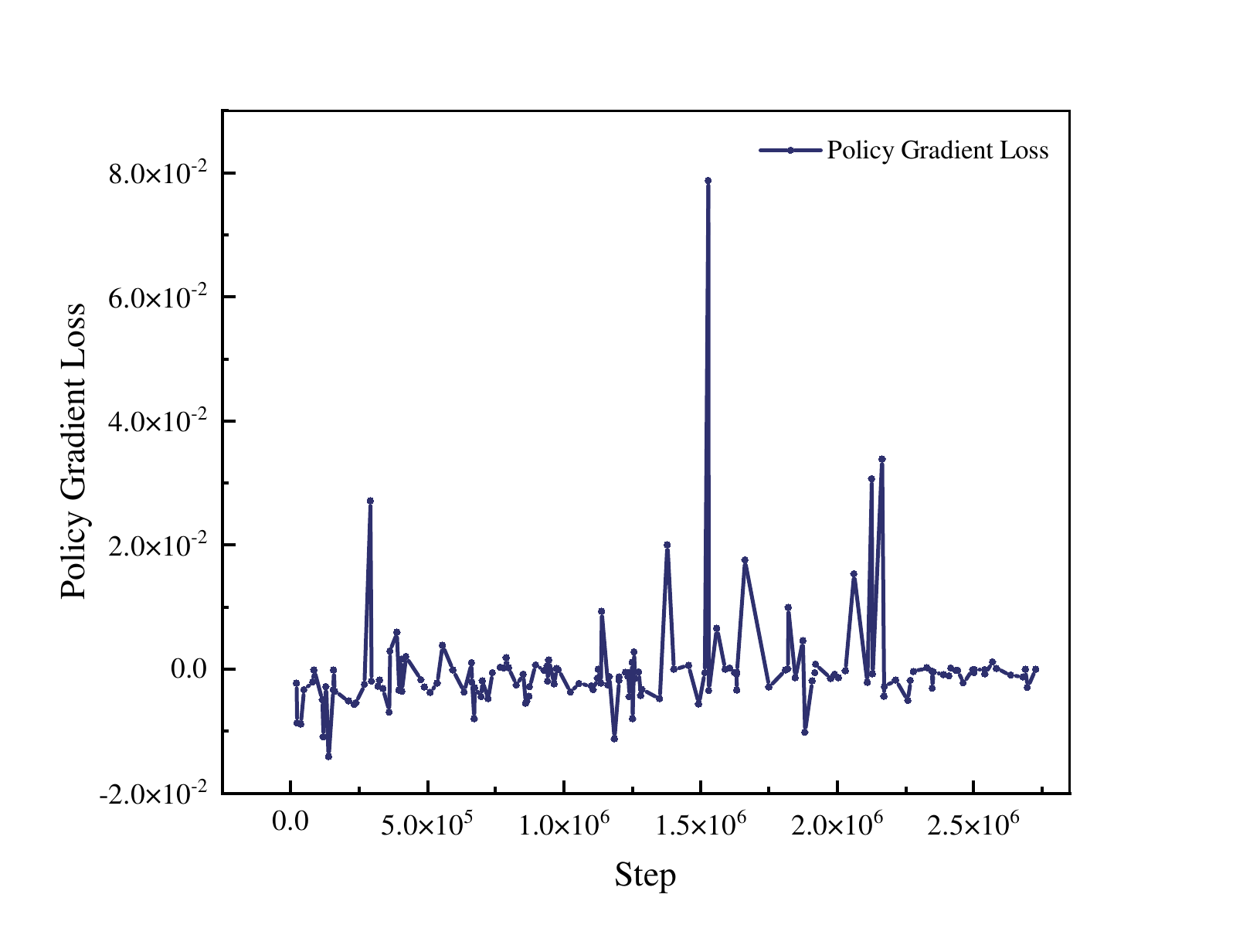}
			\label{fig:pg_loss}
		}
		\subfigure[ Value Loss]{
			\includegraphics[width=0.3\textwidth]{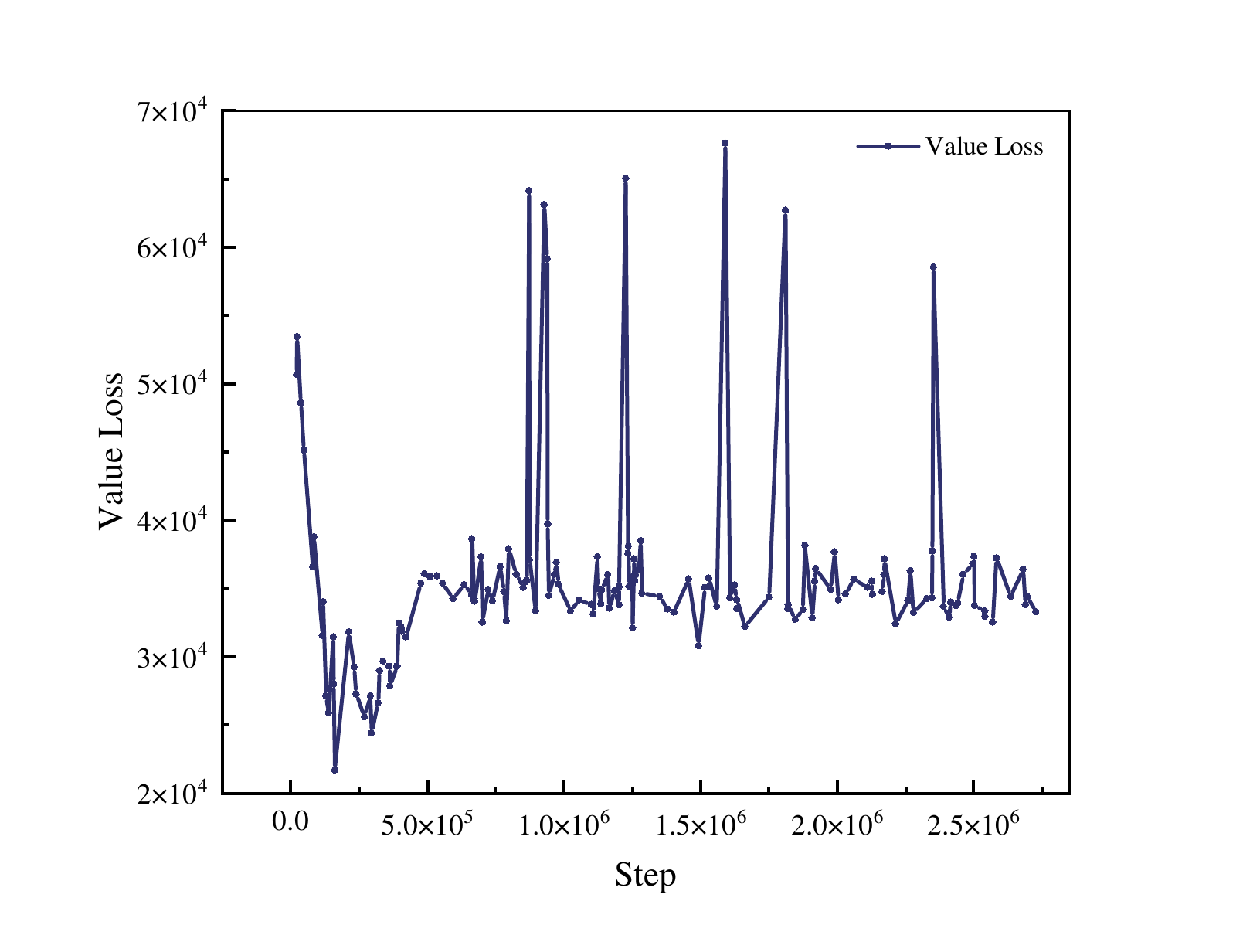}
			\label{fig:value_loss}			
		}
		\subfigure[Entropy Loss]{
			\includegraphics[width=0.3\textwidth]{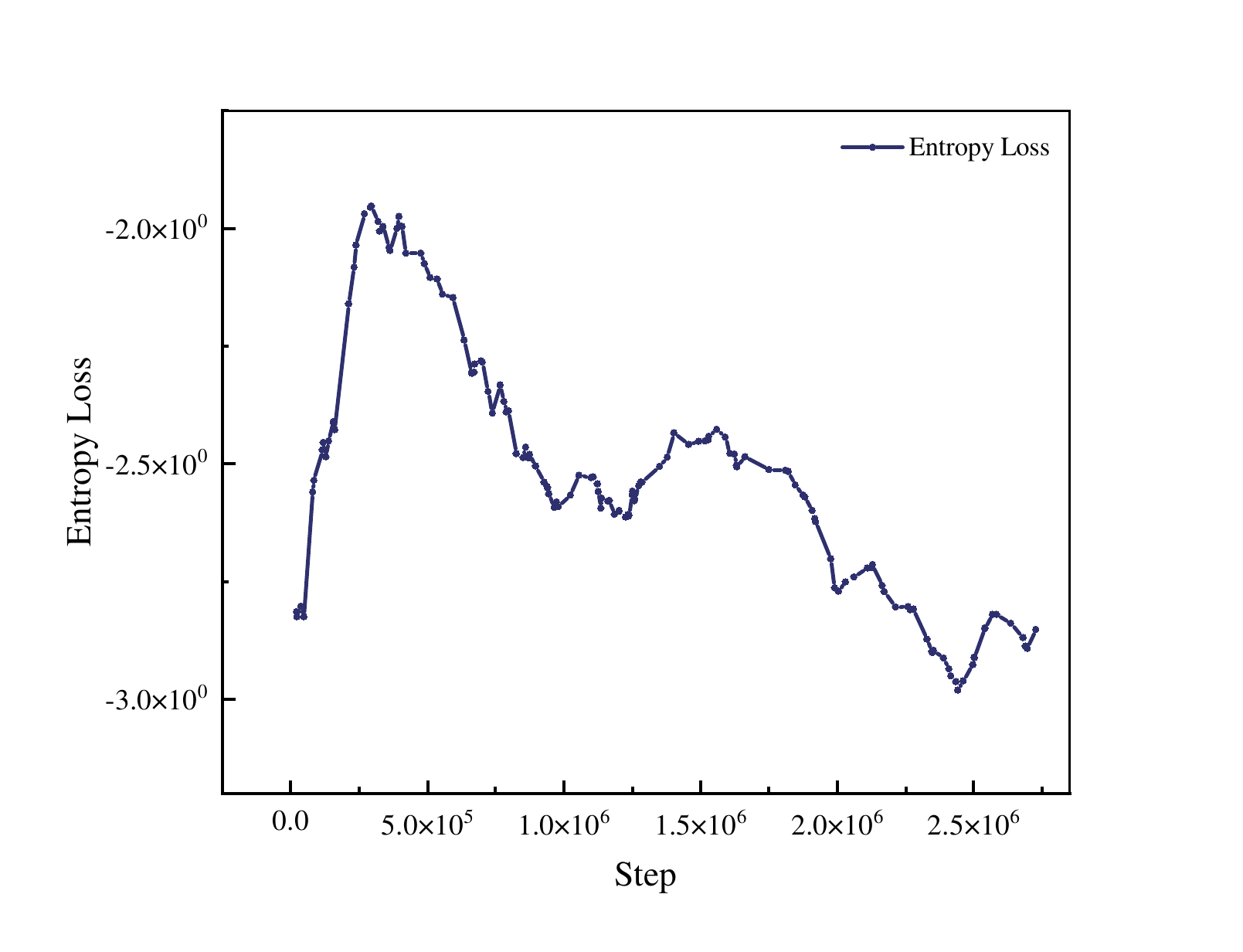}
			\label{fig:entropy_loss} 		
		}
		\caption{Results of Three Loss Functions during Training}
		\label{fig: three_loss}
	\end{figure}

Figure \ref{fig: three_loss} displays the three types of loss incurred during the training process of the PPO algorithm: policy gradient loss, value loss, and entropy loss. 
		
We applied our policy to the simulation environment and obtained the results of traction energy consumption and regenerative braking energy overlap time. The results are shown in Table \ref{table:comparision}, where we compared our proposed PPO algorithm with a baseline that takes no actions.

	\begin{table}[H]
		\caption{\label{table:comparision}Comparison Result for Our Method and No Action Baseline Result}
		\centering
		\begin{tabledata}{l^l^l^l} 
				\header Index & Our Method &No Actions\\ 
				\row Traction Energy Consumption (kWh)& 337342.6 & 378862.9\\ 
				\row Regenerative Braking Energy Overlap Time (seconds) & 7006.2 & 4734.1
			\end{tabledata}
	\end{table}

As can be seen from the table, our proposed method outperforms the baseline, reducing the traction energy consumption and increasing the regenerative braking energy overlap time. Our method achieved a reduction of 10.9\% in traction energy consumption and an increase of 47.9\% in regenerative braking energy overlap time, compared to the baseline. This demonstrates the effectiveness of our proposed algorithm in improving energy efficiency in metro systems.

\section{Conclusion}
In conclusion, this paper presents an approach to optimize energy efficiency in metro systems using the Proximal Policy Optimization (PPO) algorithm. We apply our method to a simulated metro system based on real-world data and compare its performance with a baseline method that does not consider regenerative braking energy utilization. The results demonstrate that our method outperforms the baseline by reducing traction energy consumption and increasing regenerative braking energy overlap time. Our proposed algorithm achieved a 10.9\% reduction in traction energy consumption and a 47.9\% increase in regenerative braking energy overlap time compared to the baseline. This shows the effectiveness of our proposed approach in improving energy efficiency in metro systems.
    
    \
    
Future work could explore the application of our approach in real-world metro systems and further optimize energy efficiency. Additionally, incorporating other factors such as passenger flow and train schedules could be beneficial to further improve the performance of our algorithm. Overall, the approach presented in this paper provides a promising direction for optimizing energy efficiency in metro systems, contributing to the development of sustainable transportation.

\clearpage

{\small\bibliographystyle{IEEEtran}
\bibliography{reference}}

\end{document}